\documentclass[a4paper,conference]{IEEEtran}
\IEEEoverridecommandlockouts

\usepackage{cite}
\usepackage{ntheorem}
\usepackage{algorithmic}
\usepackage{graphicx}
\usepackage{textcomp}
\usepackage{xcolor}
\usepackage{hyperref}
\usepackage{mathtools}
\usepackage{dirtytalk}
\usepackage{multirow}
\usepackage{comment}
\usepackage{amsmath} 
\usepackage{mathtools} 
\usepackage{array} 
\usepackage{xcolor} 
\usepackage{booktabs}
\usepackage{orcidlink}

\usepackage{url}
\usepackage{mwe} 
\usepackage{float}
\usepackage{gensymb}
\usepackage{amssymb}
\usepackage[version=4]{mhchem}
\usepackage{lipsum}
\usepackage{tabularx}
\usepackage{tikz}
\usepackage{pgfplots}
\pgfplotsset{compat=1.18}
\usetikzlibrary{positioning}
\usepackage{float}
\usepackage{subfigure}
\usepackage[acronym]{glossaries}
 \glsdisablehyper

\newcommand{\commentout}[1]{}

\def\BibTeX{{\rm B\kern-.05em{\sc i\kern-.025em b}\kern-.08em
    T\kern-.1667em\lower.7ex\hbox{E}\kern-.125emX}}

\begin{document}

\title{Fine-Grained HDR Image Quality Assessment\\ 
From Noticeably Distorted to Very High Fidelity}

\author{
\IEEEauthorblockN{
Mohsen Jenadeleh\orcidlink{0000-0002-8216-1195}\IEEEauthorrefmark{1},
Jon Sneyers\orcidlink{0009-0008-1697-2855}\IEEEauthorrefmark{2},
Davi Lazzarotto\IEEEauthorrefmark{3}, 
Shima Mohammadi\orcidlink{0009-0001-9352-9311}\IEEEauthorrefmark{4},
Dominik Keller\orcidlink{0000-0003-3361-7499}\IEEEauthorrefmark{5},\\
Atanas Boev\orcidlink{0000-0003-0863-4000}\IEEEauthorrefmark{6},
Rakesh Rao Ramachandra Rao\orcidlink{0000-0002-7069-1543}\IEEEauthorrefmark{5},
António Pinheiro\orcidlink{0000-0002-5968-9901}\IEEEauthorrefmark{8},
Thomas Richter\orcidlink{0000-0001-7721-0426}\IEEEauthorrefmark{9},\\
Alexander Raake\orcidlink{0000-0002-9357-1763}\IEEEauthorrefmark{7}\IEEEauthorrefmark{5},
Touradj Ebrahimi\orcidlink{0000-0002-9900-3687}\IEEEauthorrefmark{3},
João Ascenso\orcidlink{0000-0001-9902-5926}\IEEEauthorrefmark{4},
Dietmar Saupe\IEEEauthorrefmark{1}}
\IEEEauthorblockA{
\small
\IEEEauthorrefmark{1}University of Konstanz, Germany \hspace{1em}
\IEEEauthorrefmark{2}Cloudinary, Belgium \hspace{1em}
\IEEEauthorrefmark{3}EPFL, Switzerland \hspace{1em}
\IEEEauthorrefmark{4}IST-IT, Portugal \hspace{1em}
\IEEEauthorrefmark{6}Huawei, Germany \hspace{1em}\\
\IEEEauthorrefmark{5}TU Ilmenau, Germany \hspace{1em}
\IEEEauthorrefmark{7}RWTH Aachen, Germany \hspace{1em}
\IEEEauthorrefmark{8}IT-UBI, Portugal \hspace{1em}
\IEEEauthorrefmark{9}Fraunhofer IIS, Germany}
\IEEEauthorblockA{
\footnotesize
\texttt{\{mohsen.jenadeleh, dietmar.saupe\}@uni-konstanz.de, jon@cloudinary.com,}\\
\texttt{\{davi.lazzarotto, touradj.ebrahimi\}@epfl.ch, \{shima.mohammadi, joao.ascenso\}@lx.it.pt,}\\
\texttt{\{dominik.keller, rakesh-rao.ramachandra-rao\}@tu-ilmenau.de, raake@ient.rwth-aachen.de,}\\
\texttt{atanas.boev@huawei.com, antonio.pinheiro@ubi.pt, thomas.richter@iis.fraunhofer.de}
}
\thanks{Funded by the DFG (German Research Foundation) -- Project ID 496858717, titled “{JND}-based Perceptual Video Quality Analysis and Modeling”. D.S. is funded by DFG Project ID 251654672.}
}

\IEEEoverridecommandlockouts

\maketitle
\begin{abstract}
High dynamic range (HDR) and wide color gamut (WCG) technologies significantly improve color reproduction compared to standard dynamic range (SDR) and standard color gamuts, resulting in more accurate, richer, and more immersive images. However, HDR increases data demands, posing challenges for bandwidth efficiency and compression techniques. 
 Advances in compression and display technologies require more precise image quality assessment, particularly in the high-fidelity range where perceptual differences are subtle. 
 To address this gap, we introduce AIC-HDR2025, the first such HDR dataset, comprising 100 test images generated from five HDR sources, each compressed using four codecs at five compression levels.  It covers the high-fidelity range, from visible distortions to compression levels below the visually lossless threshold. 
 A subjective study was conducted using the JPEG \mbox{AIC-3} test methodology, combining plain and boosted triplet comparisons. In total, 34,560 ratings were collected from 151 participants across four fully controlled labs. The results confirm that \mbox{AIC-3}  enables precise HDR quality estimation, with 95\% confidence intervals averaging  a width of 0.27 at 1 JND. In addition, several recently proposed objective metrics were evaluated based on their correlation with subjective ratings. The dataset is publicly available\footnote{\href{https://github.com/jpeg-aic/AIC-HDR2025}{https://github.com/jpeg-aic/AIC-HDR2025}}.
 \end{abstract} 


\begin{tikzpicture}[overlay, remember picture]

\path (current page.north) node (anchor) {};

\node [below=of anchor] {%

};

\end{tikzpicture}

\section{Introduction}
\label{sec:introduction}
Advancements in imaging and display technology have significantly improved screen brightness, contrast and color accuracy. Yet standard dynamic range (SDR) imaging remains limited by its 8-bit depth, Rec.\ 709 color gamut \cite{ITU-R_BT.709}, and a reference luminance of 100 cd/m$^2$ (nits) originally defined for CRT displays.
Although modern SDR screens commonly reach peak luminance levels of 300–400 cd/m², the dynamic range and precision of the content are still limited.

High dynamic range (HDR) imaging addresses these limitations by increasing the contrast range, the bit depth (10-bit or higher), and expands the color gamut as defined in ITU-R BT.2020~\cite{ITU2020}. It employs advanced transfer functions such as the perceptual quantizer (PQ)~\cite{SMPTE2084} and hybrid log-gamma (HLG), both standardized in ITU-R BT.2100~\cite{ITU-R_BT2100-3_2025}. These advancements enable peak brightness levels up to 10,000~cd/m$^2$, significantly improving contrast, and allowing more precise color reproduction, delivering a more lifelike visual experience.  HDR technology corresponds more closely to the dynamic sensitivity of the human visual system \cite{kunkel2016perceptual, shang2023study}.

As a result, recent image compression formats, including JPEG XL~\cite{alakuijala2019jpeg}, AVIF~\cite{barman2020evaluation}, JPEG XT~\cite{artusi2019overview}, and JPEG AI~\cite{alshina2024jpeg} have integrated support for HDR imaging. 
These formats provide improved compression efficiency, color accuracy, and dynamic range handling. However, there is a notable lack of annotated HDR image quality assessment (IQA) datasets compressed using these formats. Existing datasets often rely on older HDR standards, lack proper HDR metadata, consider only one codec, or primarily contain tone-mapped representations rather than true HDR content. 
They do not sufficiently cover the high-fidelity range, particularly the range of distortion levels associated with detection probabilities below the visually lossless threshold, which is critical in HDR photography workflows where room for post-production adjustments is needed. 
This gap in HDR benchmarking limits the evaluation of codec performance, compression artifacts, and encoding strategies.

To address this, a new fine-grained HDR image dataset was designed for compatibility with the latest HDR standards. It includes five diverse HDR source images, with a total of 100 compressed versions generated using four codecs at five different bitrates.

The main contributions of this work are as follows:
\begin{itemize}
    \item Introducing the first publicly available fine-grained HDR IQA dataset, consisting of 100 test images encoded using four codecs: JPEG~XL, JPEG~AI, AVIF, and JPEG~XT.
    \item Conducting a large-scale subjective experiment and collect 34,560 ratings from 151 participants across four fully controlled laboratories.
    \item Applying the JPEG AIC-3 methodology and unified scale reconstruction to estimate precise JND scores.
    \item Benchmarking a wide range of objective IQA metrics for JND prediction.
\end{itemize}

\section{Related work}
The evaluation of subjective image quality has been extensively studied and standardized, notably by the International Telecommunication Union through ITU-T P.910 \cite{P.910} and ITU-R BT.500 \cite{BT.500}. These guidelines define widely adopted methodologies for visual quality assessment. The absolute category rating with hidden reference (ACR-HR) method involves independent quality ratings of both reference and distorted images, typically on a five-point scale.  
The double stimulus impairment scale (DSIS) is a variant in which observers compare a distorted image with an explicit reference and rate the impairment level. In paired comparisons (PC), a forced-choice paradigm is used, where observers identify the higher- or lower-quality image. The results are often analyzed using Thurstone’s Case V model to reconstruct perceptual quality scales in just noticeable difference (JND) units.

To support the development and validation of HDR IQA  metrics, a few HDR IQA datasets have been introduced. 
In \cite{Narwaria2013}, a pipeline is explored where HDR images are tone-mapped into SDR and then JPEG-compressed at different bitrates, employing tone-mapping optimization based on both PSNR and SSIM metrics.  
Another dataset was later generated \cite{mantel2014comparing} with HDR images compressed with JPEG XT at five bitrates defined with different parameters values for the compression of the SDR images and the quantization residuals. The obtained mean opinion scores (MOS) were used to benchmark the performance of a small number of quality metrics. 
Still in the context of the evaluation of JPEG XT, a later study \cite{korshunov2015subjective} encoded a set of HDR images at different bitrates with different profiles, which were evaluated using the DSIS protocol. 

A later study \cite{mikhailiuk2021consolidated} merged previous SDR and HDR datasets to create a unified dataset with more stimuli. 
The different baseline scores were obtained under varying evaluation protocols, including both ranking-based and rating-based methods. Unification was achieved through dataset alignment and re-scaling of scores using a psychometric function, resulting in the UPIQ dataset. Another recent effort to annotate a large set of HDR stimuli is the HDRC \cite{liu2024hdrc} database, produced from 80 reference images compressed with JPEG~XT and VVC and assessed with the DSIS methodology. 
  
However, the difference between consecutive distortion levels is typically large in these datasets, and the aforementioned 
methodologies are limited in reliably capturing subtle differences between stimuli.
To address this, the ISO/IEC 29170-2 standard, also known as JPEG AIC-2 \cite{AIC2}, defines a subjective methodology using a flicker test to distinguish visually lossless from visually lossy compressed images.

Recent advancements in subjective quality assessment have led to the introduction of a fine-grained evaluation methodology by JPEG AIC-3 \cite{ISOJPEGAIC2025,testolina2025fine}, building on prior research in JND-based perceptual quality scaling \cite{men2021subjective}. This methodology enhances the detection of subtle artifacts and enables the reconstruction of subjective quality scores in fractional JND units, leveraging established principles from psychophysics. It comprises two evaluation methods: plain triplet comparisons (PTC) and boosted triplet comparisons (BTC).

In the PTC method, compressed versions of a source image are displayed side by side, with observers toggling both images simultaneously with their source to evaluate perceptual differences. The BTC method builds on this approach by applying boosting techniques, such as zooming and artifact amplification, to make distortions more perceptible. During BTC assessments, two boosted test images are shown side by side, alternating (flickering) at 10 Hz with the their source image. Observers identify the image with the most pronounced flicker effect, indicating lower perceived quality.

To estimate quality scales in fractional JND units, a unified functional reconstruction approach based on the Thurstone Case V model was employed, integrating responses from both methods, PTC and BTC. This combined approach facilitates a more precise quantification of perceptual quality degradation in the compressed images.

\section{HDR image dataset}
\begin{figure*}[t!]
\centering
\begin{minipage}{0.95\textwidth}
\centering
    \subfigure[alps-sunset]{
    \begin{minipage}[t]{0.19\textwidth}
    \centering
    \includegraphics[width=\linewidth]{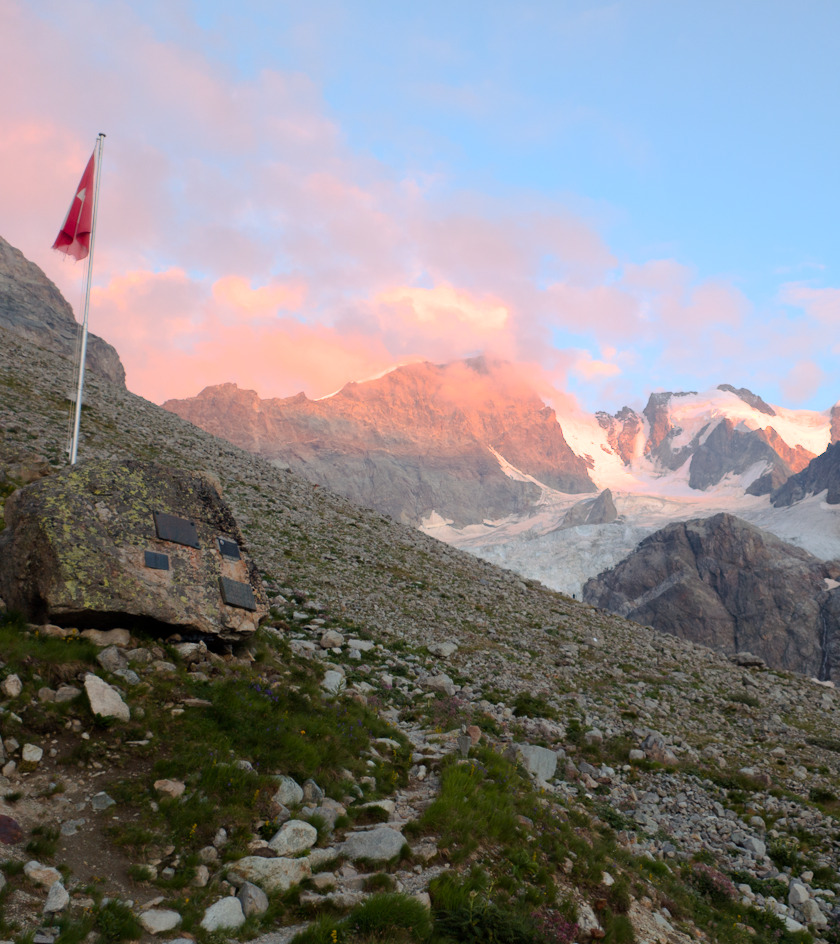}
    \end{minipage}}%
    \subfigure[building]{
    \begin{minipage}[t]{0.19\textwidth}
    \centering
    \includegraphics[width=\linewidth]{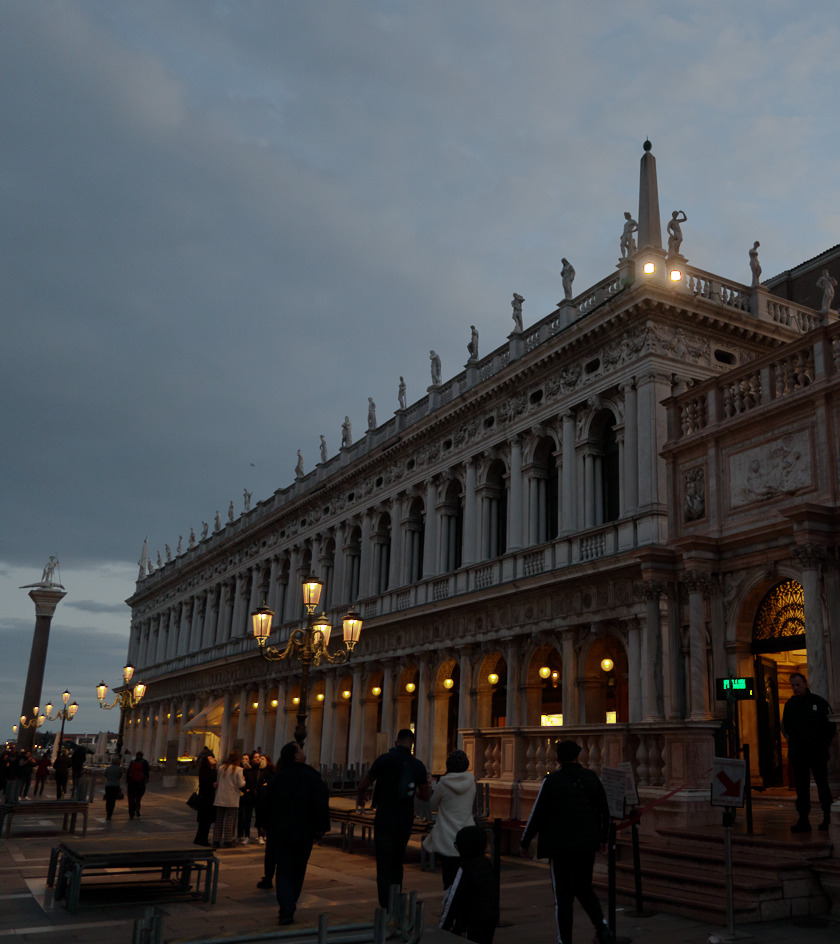}
    \end{minipage}}%
    \subfigure[flower]{
    \begin{minipage}[t]{0.19\textwidth}
    \centering
    \includegraphics[width=\linewidth]{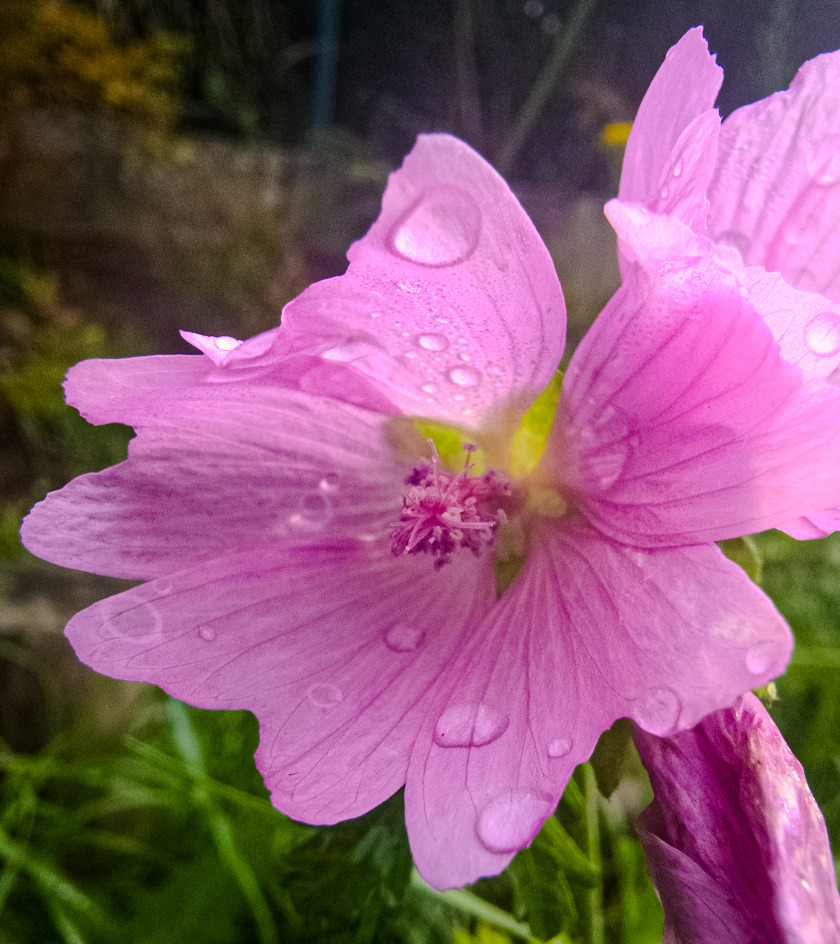}
    \end{minipage}}%
    \subfigure[p29-crop]{
    \begin{minipage}[t]{0.19\textwidth}
    \centering
    \includegraphics[width=\linewidth]{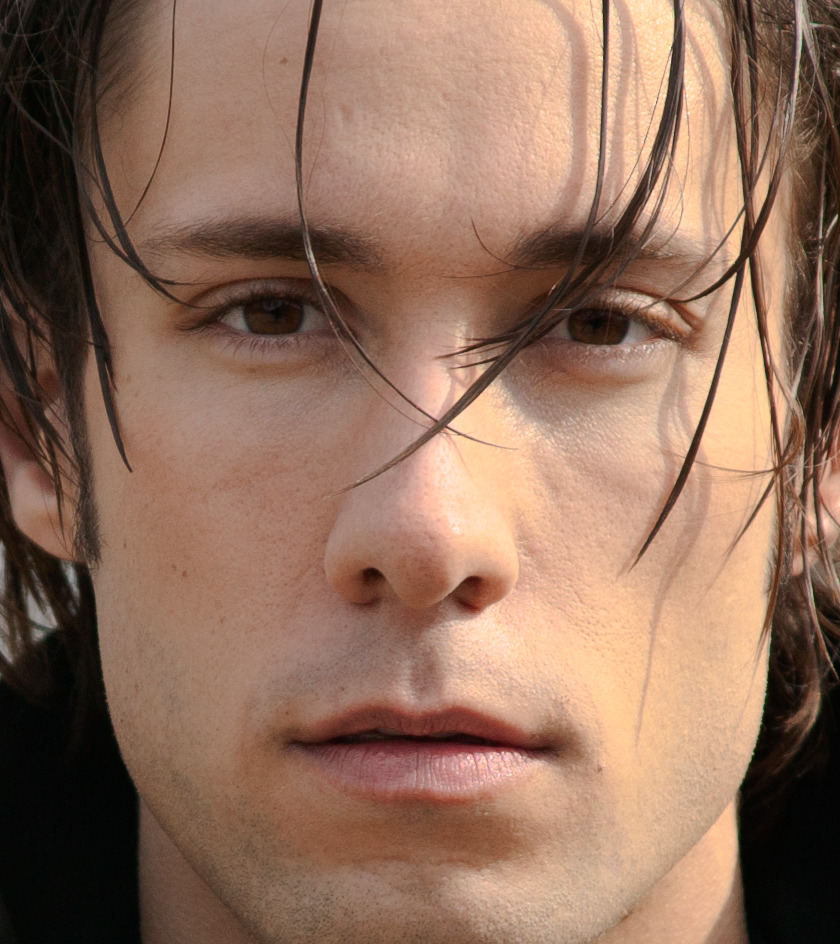}
    \end{minipage}}%
    \subfigure[room1]{
    \begin{minipage}[t]{0.19\textwidth}
    \centering
    \includegraphics[width=\linewidth]{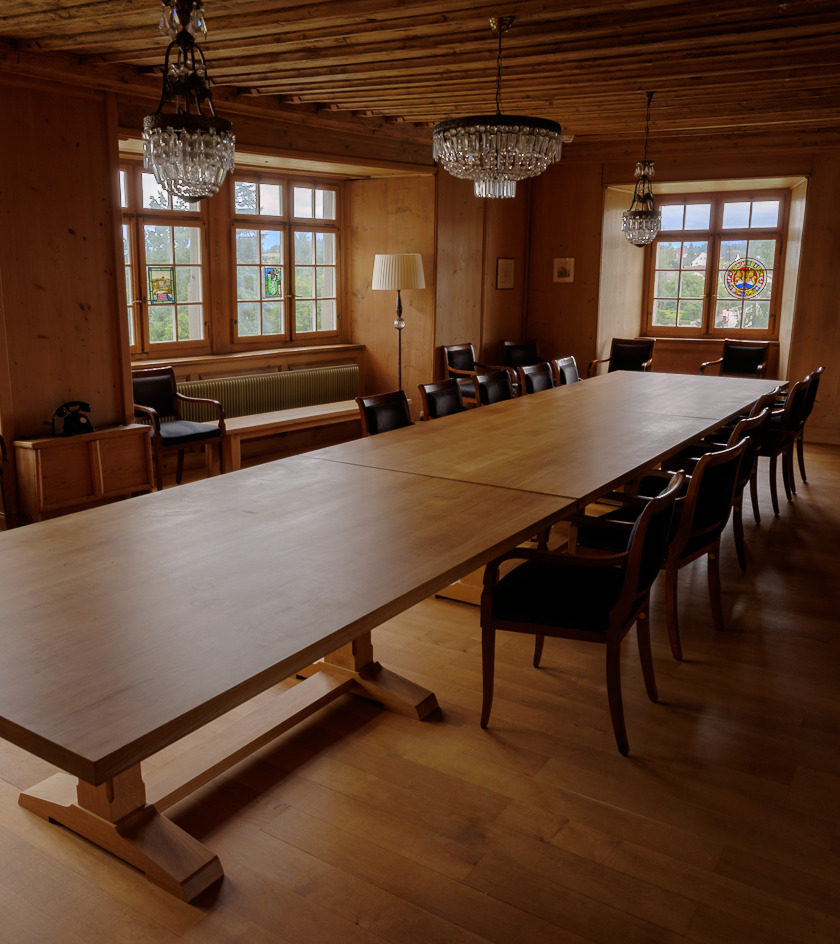}
    \end{minipage}}%
    \vspace{-5pt}
       \caption{Source HDR images used in this study. All images have a resolution of $840 \times 944$ pixels. The previews shown here are tone-mapped to SDR and a standard color gamut. To view the HDR source images with accurate color, please visit  \href{https://cloudinary.com/labs/aic-3-and-hdr}{https://cloudinary.com/labs/aic-3-and-hdr}.}
    \vspace{-10pt}
    \label{fig:sources}
\end{minipage}
\end{figure*}
\subsection{Source images}
To select five source images for our HDR study, 23 HDR images with varying resolutions were reviewed. 
The selection process aimed to ensure diversity in content. The final set includes five images from distinct categories: Landscapes (alps-sunset), Night Scene (building), Nature (flower), Portrait (p29-crop), and Architecture (room1). Fig.~\ref{fig:sources} presents the selected source HDR images. 
The source images were created from camera raw files using Adobe Lightroom, applying best practices as discussed in
\cite{itu-bt2390-8} and \cite {hdr_explained}: most of the image content remains within the SDR range, with the brightest regions approximately +2 stops above SDR white; in some images, small details going up to +4 stops.
The images were rescaled if needed and cropped to a resolution of $840 \times 944$ pixels (to fit two zoomed images on a test display), and are represented as 10-bit RGB images in the Rec. 2100 color space with the PQ transfer function. 
For display in a web browser with HDR support, all images were converted to PNG files tagged with an ICC v4.4 color profile (ISO/DIS 15076-1:2024) containing a cicpTag specifying the Rec. 2100 PQ color space.

\subsection{Encoder recipes}
Although  codec comparison is not the objective of this work, experts from each codec team were consulted to review our encoding settings and provide feedback. Their recommendations were incorporated to ensure appropriate encoding configurations. The following encoding recipes were used: 

\paragraph{JPEG XT}
Reference software libjpeg~v1.70 was used with command line {\footnotesize \verb|jpeg -h -qt 3 -oz -R 2 -q Q|}.
These flags correspond to optimized Huffman coding,
quantization tables suggested by ImageMagick (perceptually optimized),
optimized trellis quantization, 2 refinement bits to increase the precision of the DCT coefficients, and no chroma subsampling (i.e., 4:4:4). The parameter \verb|-q| selects the quantization strength.   
While JPEG XT offers scalable two-layer coding, the second residual layer is omitted in this experiment to avoid rate allocation between layers. Instead, the bit depth of the first layer is extended.
\paragraph{JPEG XL}
Reference software libjxl~v0.11.1 was used with command line {\footnotesize \verb|cjxl -x color_space=Rec2100PQ -q Q|}.
This tags the input PPM with the correct color space; the parameter \verb|-q| selects the quality.

\paragraph{JPEG AI}
The JPEG AI verification model software (v7.0) was used in 10-bit YUV 4:4:4 mode:
\[
\parbox{0.99\linewidth}{\footnotesize \ttfamily
python -m src.reco.scripts.eval --use\_yuv 1\\
--cfg ./cfg/tools\_on.json ./cfg/oper\_point/hop.json \\ ./cfg/BRM/regen\_list.json \\
--coding\_type enc\_dec -target\_bpps [BPP*100] 
}
\]
Executing this command enables all JPEG AI coding tools, generating a bitstream with a bitrate approximating the target BPP and the corresponding decoded image. Note that the current  version of the JPEG AI VM defines BT.709 as the color space used for coding (does not support wide color gamuts) and was trained only for SDR (BT.709) image data and thus is not expected to have optimal compression performance when used on HDR images. 
\paragraph{AVIF}
The reference implementation libavif v1.1.1 with libaom 3.11.0 was used, with the following command line:
{\footnotesize \verb|avifenc --autotiling -d 10 --cicp 9/16/9 -q Q|}.

The encoding recipes and further details of the parameters for all codecs will be made publicly available with the dataset.
\subsection{Target bitrates selection}
 
For each source image, a set of target bitrates was manually selected based on visual inspection of the encoded images.
The goal was to span a range of distortion levels up to 
3~JND, with roughly evenly spaced distortion levels.

For all codecs, except JPEG AI, 
the bitrate for each test image was matched to a target value using binary search by adjusting the \verb|Q| parameter. JPEG AI utilized its internal bitrate matcher, which combines model selection, linear interpolation, and binary search to find the $\beta$ parameter that matches the output rate. Moreover, due to its legacy nature, JPEG XT required higher bitrates; they were set to twice the target bitrate used for the other codecs, plus a source-dependent constant of 0.1 to 0.3 bits per pixel. This adjustment accounts for the lower coding efficiency of JPEG~XT, ensuring a comparable visual quality range. Across all source images and codecs, the actual bitrates closely matched their targets, with an average absolute difference of 2\% and a maximum difference of 9\%.

The final \verb|Q| settings that were used to encode the images were in the following ranges. For JPEG~XT the lowest \verb|Q| setting (highest distortion) was between 15 and 33 (depending on the source image), while the highest \verb|Q| setting was between 79 and 88.
For JPEG~XL, the lowest quality setting was between 12 and 24, while the highest 
was between 79 and 88.
For AVIF, the lowest quality setting was between 45 and 61, while the highest 
was between 76 and 87. 
Note that these \verb|Q| settings use an encoder-specific quality scale that cannot meaningfully be compared between codecs.

\section{Subjective evaluation}
A new dataset, HDR-AIC2025, was constructed by compressing each of the five source images at five different bitrate levels using four different codecs, yielding a total of 100 compressed images. To estimate the perceived quality in JND units, we adopted the subjective evaluation methodology of JPEG~\mbox{AIC-3}~\cite{ISOJPEGAIC2025,testolina2025fine}, which is specifically designed to capture fine-grained quality differences in distorted images.

This methodology evaluates boosted and plain triplet comparisons. In BTC, the test images were zoomed by a factor of 2 and test images alternate with their source at 10 Hz to induce a flicker effect.  Observers choose the image with the stronger flicker or select ``not sure'' if undecided. As shown in \cite{jenadeleh2023relaxed}, not forcing a choice helps to reduce mental load without affecting function homogeneity. In PTC, a toggle button allows observers to switch between the compressed images and their source (in-place), with at least one toggle required before submitting a response.
\subsection{Batch generation}
Triplets for the BTC and the PTC methods were generated following the procedure described in \cite{testolina2025fine}. Each triplet $(I_i,I_0,I_j)$ consists of a source image $I_0$ as pivot and two compressed versions $I_i$ and $I_j$.

For each choice of a source image and a codec, there are five distinct distortion levels plus the source image itself. Two types of triplet comparisons were generated:
\begin{itemize}
    \item Same-codec: Triplets consisting of images compressed with the same codec at different distortion levels.
    \item Cross-codec: Triplets with images compressed by different codecs, used to align quality scales.
\end{itemize}

Each source image was associated with 30 same-codec BTC and 30 same-codec PTC triplets per codec, totaling 120 same-codec triplets across four codecs. An additional 24 cross-codec triplets were included per image. This results in 16.7\% cross-codec triplets relative to the total, slightly lower than the 20\% recommended in the JPEG AIC-3 standard \cite{ISOJPEGAIC2025}. Thus, each source image was associated with a total of 144 triplets for BTC experiments and 144 triplets for PTC experiments. 
Consequently, the total number of triplet comparisons for five source images was 720 for BTC and 720 for PTC.

To facilitate the experimental procedure, the 720 triplet questions for each type of experiment were divided into 6 batches of 120 questions each. The triplets were arranged so that if a triplet $(I_i,I_0,I_j)$ is included in a batch, its mirrored counterpart $(I_j, I_0, I_i)$ was also included in the same batch to allow the assessment of the consistency of the responses.

\subsection{Laboratory-based subjective quality assessment}

The web interfaces developed by JPEG AIC-3 for the PTC and BTC test methodologies \cite{testolina2025fine} were used and customized for the subjective experiment, enabling viewing of the triplets on 4K screens. Google Chrome version 134.x on macOS version 15.x was used. 
In both BTC and PTC experiments, images were shown at a 1:1 pixel-to-native-display resolution on a 4K HDR screen. BTC used the full display, while PTC used only the central 2K region (about one-quarter). 

The viewing distance for the PTC experiment was set to 3.1 times the stimulus height, and for the BTC experiment, it was set to 1.5 times the stimulus height, following the ITU-R BT.2246-8 recommendation \cite{ITU-R_BT2246-8_2023}. The screen center was aligned with the participant's eye level, and the screen was positioned perpendicular to the line of sight to minimize glare and distortion. An external keyboard and mouse were provided for participants' convenience.

Following the recommendation ITU-R BT.2100-3 for critical viewing of HDR content \cite{ITU-R_BT2100-3_2025}, the incident ambient light was calibrated to 5 lux in a fully controlled lab environment. The experiments, conducted in four fully controlled laboratories, located in three countries, each collected a quarter of the total number of responses. In three laboratories, MacBook Pro models (2021 or later, 16-inch) featuring Liquid Retina XDR displays were used, configured to the native resolution of $3456 \times 2234$ pixels  at a refresh rate of 60 Hz. The displays used an HDR P3 color profile with SDR white calibrated to D65 at 100 cd/m$^2$ and a peak brightness set to 1000 cd/m$^2$ (i.e., 3.32 stops of HDR headroom). In one laboratory, a Sony BVM-HX310 display was configured to its native resolution of $3840 \times 2160$ pixels and connected to a Mac mini via HDMI 2, with the refresh rate set to 30 Hz to ensure sufficient bandwidth for the BTC interface and to avoid chroma subsampling. The same setting for SDR white and peak brightness was used.

For both the PTC and BTC experiments, 24 responses per triplet were collected. Separate groups of participants were recruited for each experiment and each participant was allowed to complete a maximum of two batches (each consisting of 120 questions) within a single experiment. This limitation was enforced to maintain the total duration of the experiment according to the guidelines recommended by ITU-T BT.500 and to minimize potential fatigue effects on response accuracy. Most participants (137) completed two batches, while some (14) completed only one. The average duration per batch was 20.1 minutes for PTC and 10.7 minutes for BTC. 
Between the two batches, there was a mandatory three-minute break.  Participants were allowed to take longer breaks if desired.

Table~\ref{tbl:exp_data} summarizes the statistics of the collected triplet responses, including accuracy and average response time per triplet. As depicted in the table, response accuracy increases with greater differences in distortion levels between the test images presented in each triplet, as expected.
\begin{table}[t!]
\centering
\scriptsize 
\setlength{\tabcolsep}{3pt}
\renewcommand{\arraystretch}{1.05}
\caption{Summary of experimental data}
\vspace{-5pt}
\label{tab:summary_all}
\begin{tabular}{llccccc}
\toprule
Type & Measure & \multicolumn{5}{c}{Distortion level difference} \\
\cmidrule(lr){3-7}
 & & 1 & 2 & 3 & 4 & 5 \\
\midrule
\multirow{2}{*}{BTC}
  & Avg. time (s) & 5.7 & 4.7 & 4.1 & 3.5 & 3.1 \\
  & Rsp.\ ratio correct   & 0.637 &    0.865 &    0.942 &   0.977 &    0.985 \\
  & Rsp.\ ratio not sure  & 0.316 &    0.118 &    0.044 &   0.012 &    0.001 \\
  & Rsp.\ ratio incorrect & 0.047 &    0.016 &    0.015 &   0.012 &    0.014 \\
\midrule
\multirow{2}{*}{PTC}
  & Avg. time (s) & 10.8 & 9.4 & 8.1 & 6.8 & 5.9 \\
  & Rsp.\ ratio correct   & 0.560 &    0.720 &    0.830 &    0.925 &    0.977 \\
  & Rsp.\ ratio not sure  & 0.359 &    0.228 &    0.143 &    0.051 &    0.017 \\
  & Rsp.\ ratio incorrect & 0.080 &    0.052 &    0.028 &    0.024 &    0.006 \\
\midrule
 & Response counts & Left & Not sure & Right & Skip & Total \\
\midrule
BTC &  & 7,838 & 2,405 & 7,019 & 18 & 17,280 \\
PTC &  & 6,708 & 3,547 & 6,889 & 136 & 17,280 \\
\bottomrule
\end{tabular}
\label{tbl:exp_data}
\vspace{-10pt}
\end{table}

All experimental procedures were approved by the relevant institutional ethics committees.

\section{Experimental results}
\commentout{
\begin{figure}[t!]
\centering
\includegraphics[width=0.4\linewidth]{figures/fig_ac_btc.png}%
\includegraphics[width=0.4\linewidth]{figures/fig_ac_ptc.png}%
\vspace{-5pt}
\caption{Scatter plots of batch accuracy and consistency for the raw BTC and PTC data. The mean of accuracy and consistency is thresholded at 0.7. }
\label{fig_accuracy_consistecy}
 \end{figure}
 }

\begin{figure*}[ht!]
\includegraphics[width=\linewidth]{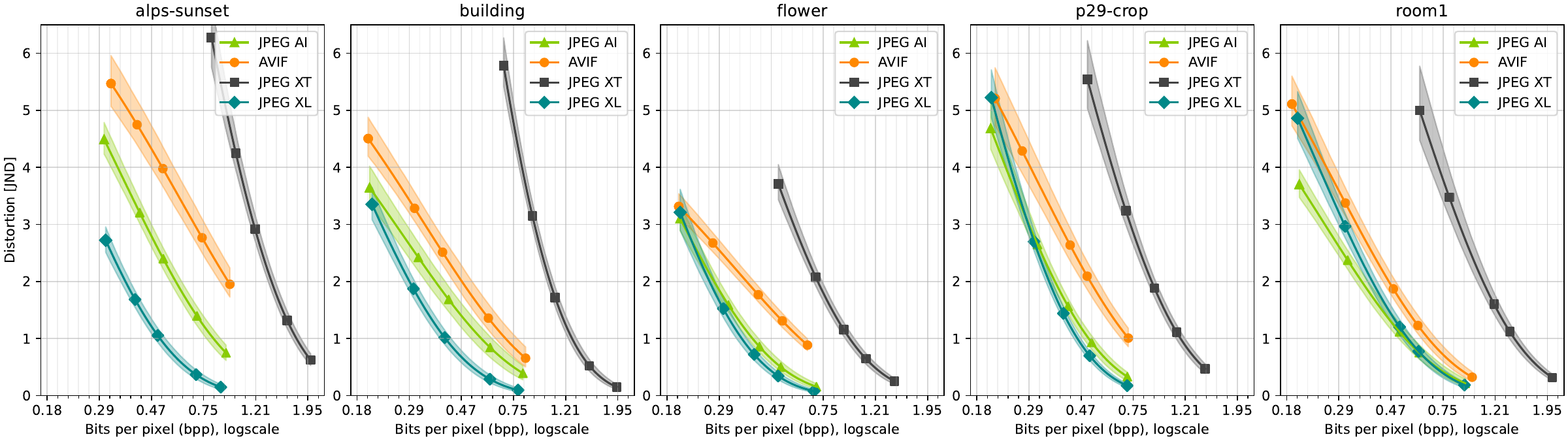}%
\vspace{-15pt}
\caption{Perceived impairment in JND units for source images and codecs. The shaded regions indicate 95\% confidence intervals.}
\label{fig:bitrate_distortion_plots}
\end{figure*}
\begin{figure*}
\centering
\includegraphics[width=0.96\linewidth]{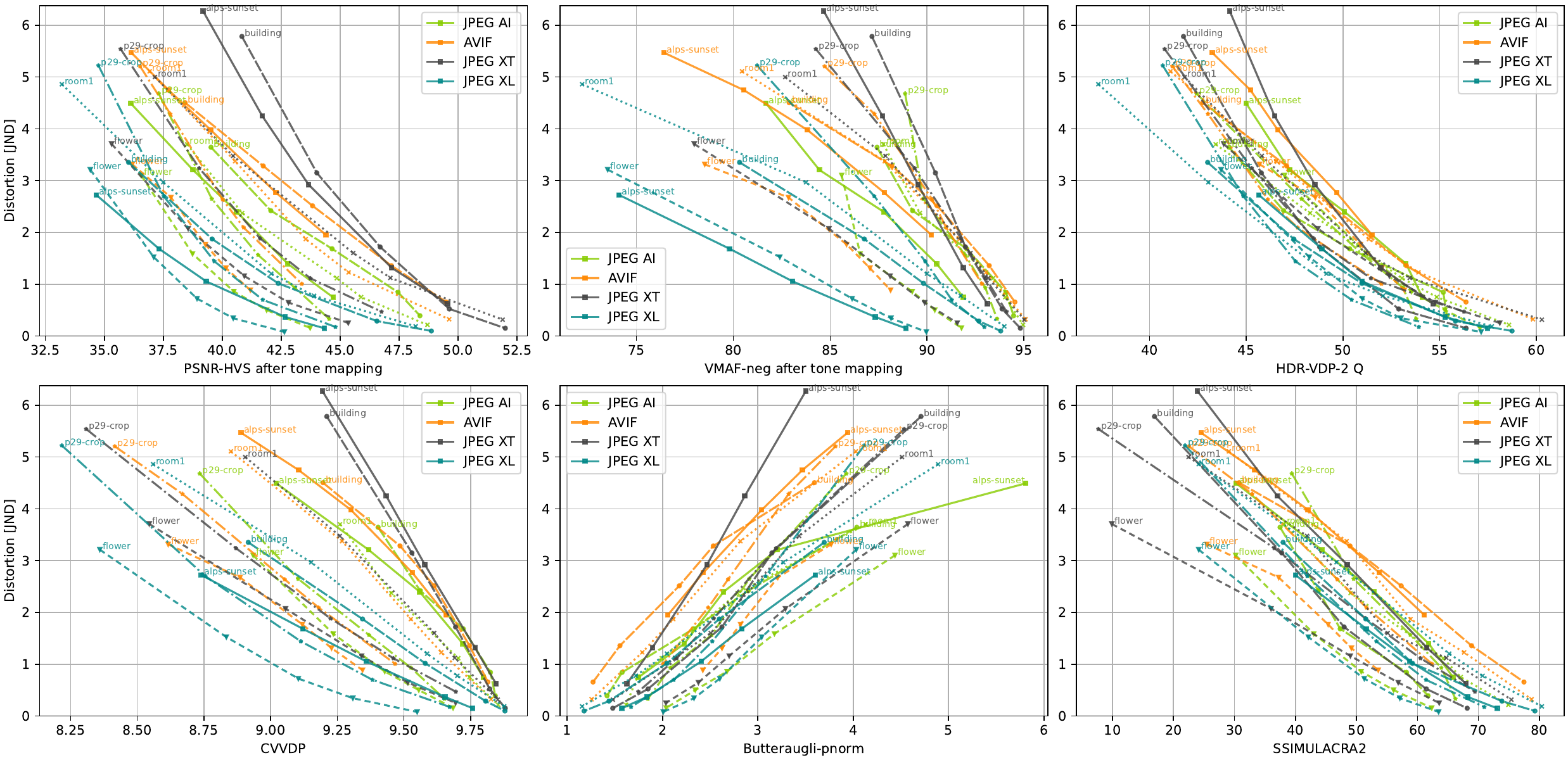}%
\vspace{-10pt}
\caption{Correlation plots between metric scores and estimated JND values, with colors denoting codecs and line styles indicating source images.}
\label{fig:metric_correlation_plots}
\vspace{-10pt}
\end{figure*}
\subsection{Data cleansing}
The collected responses for a batch of questions may be unreliable due to inattention, fatigue, or other factors. In data cleansing such instances of batches were identified, adopting a criterion from JPEG AIC-3~\cite{ISOJPEGAIC2025}.  Each batch instance, consisting of 120 responses, was assigned an accuracy and a consistency in [0,1] as follows.
\begin{itemize}
    \item \textit{Accuracy}. Considering only same-codec comparisons, a response is correct if the lower-bitrate image is judged more distorted; ``not sure'' counts as half correct. Accuracy is the distortion-weighted ratio of correct responses.  
    \item \textit{Consistency}. The questions occur in symmetric pairs by design. For each pair, a score of 1 is assigned if the two responses are consistent, that is, both are correct, incorrect, or ‘not sure’. A score of 0.375 is given if one response is ‘not sure’ and the other is not, and 0 otherwise, i.e., if both responses are either ‘left’ or ‘right’. Consistency is defined as the weighted mean of the scores per question pair, with the same weights as for accuracy.
\end{itemize}
Out of the 144 batches per experiment, those with an average accuracy and consistency below 0.7 were excluded from further analysis, resulting in 142 BTC and 134 PTC batches retained for scale reconstruction.
\subsection{Unified functional scale reconstruction}
In this contribution the JPEG AIC-3 modeling process \cite{ISOJPEGAIC2025} was applied to construct impairment scale values for the compressed HDR source images, briefly outlined as follows. For each source image the corresponding data from BTC and PTC is combined for a unified model given in the form of two functions per codec. The first is an exponential rate-distortion (RD) function $d(r) = \alpha e^{-\beta r}$, that maps bitrate $r$ to perceived distortion $d$ in JND units corresponding to the plain triplet comparison task. The second function nonlinearly maps these distortions $d(r)$ to the perceived distortion corresponding to the boosted triplet comparison task by means of a quadratic boosting transform, $h(d) = \gamma_1 d + \gamma_2 d^2$. Thus, the boosted version of a compressed image at bitrate $r$ will have a perceived distortion of $h(d(r))$. 

For each source image four codecs were used and thus a total of 16 parameters of type $\alpha, \beta, \gamma_1, \gamma_2$ were estimated.
For any assignment of values to these parameters, the functions determine distortions that are used in the classic Thurstonian Case V model to derive probabilities for each response of our dataset. The responses 'not sure' are split half and half between 'left' and 'right'. Finally, the optimal model parameters are determined by maximum likelihood estimation.

Fig.~\ref{fig:bitrate_distortion_plots} shows the RD curves for all source images and codecs, with shaded areas indicating 95\% confidence intervals (CI). These CIs were computed by bootstrapping the cleansed dataset 1,000 times: for each iteration, triplet questions were resampled with replacement, RD curves were reconstructed, and distortion values were evaluated at 100 equally spaced bitrates. The 95\% CI was then calculated for each bitrate.

The estimated RD curves exhibit narrow CI, averaging 0.27 at 1 JND. The dataset spans a broad quality range: from imperceptible differences near 0 JND to obvious distortions around 3 JNDs, and up to 5 JNDs for some codecs and images. These differences often occur in small regions, e.g. in bright highlights, smooth gradients such as skies, or subtle texture such as clouds and skin. The performance of codecs is influenced by their internal color space, making direct comparisons between coding technologies unreliable. In this case, it is plausible that JPEG~XL has the best overall performance because of its use of the perceptually motivated XYB color space, as opposed to the YCbCr space used in the other codecs.
\section{Benchmarking IQA metrics}
Fig.~\ref{fig:metric_correlation_plots} provides a detailed visualization of how well different metrics capture perceived quality. It shows the relationship between the predicted scores of six representative metrics and the estimated JND scores. These metrics were chosen to reflect a diverse set of methodologies—including both HDR-specific and conventional models—and span a range of performance characteristics. Each line in the figure corresponds to a specific combination of source image and codec, offering insight into the consistency of each metric across different content and distortion types. 

Table~\ref{tbl:correlation} summarizes the Pearson (PLCC) and Spearman (SRCC) correlations between various objective quality metrics and the estimated perceived distortion, reflecting each metric’s overall performance across the dataset.
To assess the statistical significance of the reported SRCC differences, we applied the Meng–Rosenthal–Rubin test for dependent correlations~\cite{meng1992comparing}. The results show that HDR-VDP-2 significantly outperforms all other metrics when evaluating overall correlation.

For metrics that are not specifically developed for HDR images, results are reported for both the original images in the Rec.~2100 PQ color space and for tone-mapped versions of the same images. The tone mapping follows the 
approach outlined in Section 5.4 of ITU-R BT.2390-8~\cite{itu-bt2390-8}, followed by a gamut mapping to the sRGB color space. 
Some metrics perform better on tone-mapped images, possibly because they misinterpret PQ-encoded HDR as sRGB, reducing contrast and color saturation. Tone mapping helps correct this.

As the 16-inch MacBook Pro was used in three out of the four laboratories involved in the study, the CVVDP metric~\cite{cvvdp} was configured to match this display. The configuration was as follows:
{\small ``ColorVideoVDP v0.4.2, 56.55 [pix/deg], Lpeak = 1000, Lblack = 0.001, Lrefl = 0.007958 [cd/m${}^2$], (custom-display: custom\_hdr\_pq)''}. 
The HDR-VDP metrics~\cite{hdr-vdp-2, hdr-vdp-3} were likewise configured for a viewing distance of 56.55 pixels per degree, with the additional option {\small \{'rgb\_display', 'led-lcd-wcg'\}} enabled. 

In addition to the overall correlation values, Table~\ref{tbl:correlation} also reports per-codec and per-source correlation values, i.e., correlations computed on a subset of the data corresponding to a single codec or a single source, and then averaged over all codecs or sources. These values are generally higher, as they are less affected by systematic biases associated with specific codecs or image content. Although narrower in scope, per-codec correlation suffices for targeted applications, e.g., bitrate allocation—where intra-codec consistency is more critical than cross-codec generalization. Similarly, per-source correlation may be sufficient for comparing codec performance.
\begin{table}[t!]
\centering
\footnotesize
\setlength{\tabcolsep}{2.5pt}
\renewcommand{\arraystretch}{1.2}
\caption{Metric correlation coefficients PLCC and SRCC}
\label{tab:metric_correlations}
\vspace{-7pt}
\begin{tabular}{lcc|cc|cc}
\toprule
\vspace{-2pt}
& \multicolumn{2}{c|}{Overall} & \multicolumn{2}{c|}{Per-codec}& \multicolumn{2}{c}{Per-source}\\
Metric & PLCC & SRCC & PLCC & SRCC & PLCC & SRCC\\
\midrule
PSNR-Y (PQ) & 0.587 & -0.596 & 0.741 & -0.792 & 0.730 & -0.733 \\
PSNR-Y after TM & 0.612 & -0.611 & 0.698 & -0.701 & 0.813 & -0.816 \\
SSIM \cite{ssim}& 0.674 & -0.672 & 0.720 & -0.713 & 0.876 & -0.889 \\
SSIM after TM & 0.544 & -0.533 & 0.635 & -0.638 & 0.770 & -0.767 \\
MS-SSIM (PQ) \cite{ms-ssim} & 0.757 & -0.731 & 0.764 & -0.749 & \underline{0.918} & -0.907 \\
MS-SSIM after TM & 0.679 & -0.669 & 0.709 & -0.707 & 0.905 & -0.907 \\
PSNR-HVS (PQ) \cite{psnr-hvs} & 0.520 & -0.526 & 0.771 & -0.798 & 0.635 & -0.639 \\
PSNR-HVS after TM & 0.731 & -0.748 & 0.824 & -0.831 & 0.829 & -0.861 \\
IW-PSNR (PQ) \cite{iw-ssim} & 0.658 & -0.670 & 0.826 & -0.875 & 0.728 & -0.734 \\
IW-PSNR after TM & 0.805 & -0.817 & 0.853 & -0.859 & 0.900 & \underline{-0.914} \\
IW-SSIM (PQ) \cite{iw-ssim} & 0.647 & -0.658 & 0.775 & -0.806 & 0.773 & -0.780 \\
IW-SSIM after TM & 0.676 & -0.680 & 0.709 & -0.704 & 0.885 & -0.892 \\
VMAF (PQ) \cite{vmaf} & 0.584 & -0.585 & 0.740 & -0.793 & 0.688 & -0.691 \\
VMAF after TM & 0.678 & -0.680 & 0.760 & -0.784 & 0.801 & -0.791 \\
VMAF-neg (PQ) \cite{vmaf-neg} & 0.599 & -0.604 & 0.760 & -0.816 & 0.714 & -0.718 \\
VMAF-neg after TM & 0.713 & -0.721 & 0.786 & -0.808 & 0.834 & -0.837 \\
\midrule
LGFM \cite{LGFM} & 0.462 & -0.421 & 0.542 & -0.536 & 0.596 & -0.747 \\
PU2-PSNR \cite{PU21} & 0.622 & -0.621 & 0.723 & -0.734 & 0.798 & -0.792 \\
PU2-IW-PSNR \cite{PU21,iw-ssim} & \underline{0.813} & \underline{-0.828} & \underline{0.895} & \underline{-0.908} & 0.864 & -0.873 \\
PU2-SSIM \cite{PU21} & 0.677 & -0.676 & 0.720 & -0.725 & 0.878 & -0.889 \\
PU2-MS-SSIM & 0.770 & -0.743 & 0.773 & -0.758 & \underline{0.922} & \underline{-0.909} \\
PU2-IW-SSIM \cite{PU21,iw-ssim} & 0.757 & -0.768 & 0.859 & -0.882 & 0.819 & -0.837 \\
HDR-VDP-2 \cite{hdr-vdp-2} & \underline{\textbf{0.936}} & \underline{\textbf{-0.946}} & \underline{\textbf{0.963}} & \underline{\textbf{-0.965}} & \underline{0.955} & \underline{-0.949} \\
HDR-VDP-3 (detect) \cite{hdr-vdp-3} & 0.742 & -0.744 & 0.786 & -0.783 & 0.889 & -0.899 \\
CVVDP \cite{cvvdp} & 0.736 & -0.753 & 0.782 & -0.784 & 0.883 & -0.900 \\
Butteraugli-max \cite{butteraugli} & \underline{0.836} & \underline{0.845} & \underline{0.878} & \underline{0.889} & 0.888 & 0.881 \\
Butteraugli-pnorm \cite{butteraugli} & \underline{0.882} & \underline{0.886} & \underline{0.910} & \underline{0.901} & \underline{0.939} & \underline{0.947} \\
SSIMULACRA 1 \cite{ssimulacra1} & 0.546 & 0.512 & 0.569 & 0.522 & 0.819 & 0.892 \\
SSIMULACRA 2 \cite{ssimulacra2} & \underline{0.906} & \underline{-0.895} & \underline{0.907} & \underline{-0.888} & \underline{\textbf{0.968}} & \underline{\textbf{-0.958}} \\
\bottomrule
 \end{tabular}

\vspace{0.5em}
The top 5 values are \underline{underlined}, and the best is shown in \textbf{bold}.
\label{tbl:correlation}
\vspace{-15pt}
\end{table}

\section{Conclusion and future work}
In this work, we introduced a JND-based HDR image quality assessment dataset that captures fine-grained perceptual differences in image quality. 
The dataset includes 100 compressed images derived from five HDR sources, each encoded using four state-of-the-art codecs across five compression levels. Subjective quality scores were obtained via the JPEG AIC-3 methodology, combining plain and boosted triplet comparisons. A total of 34,560 ratings were collected from 151 participants across four fully controlled laboratory environments.

The results show that the AIC-3 methodology effectively captures subtle variations in HDR image quality. Confidence intervals as narrow as 0.27 around the 1 JND unit support the reconstruction of highly precise, JND-based quality scales. Several objective image quality metrics were benchmarked. 
The results highlight limitations of conventional SDR-based metrics and underscore the importance of perceptually calibrated evaluation for HDR content.

In future work, the reconstructed quality scales across different labs will be compared. 
Since one of the labs used a different display (Sony BVM-HX310) compared to the others (MacBook Pro XDR), an analysis will be conducted to determine whether there are statistically significant deviations in the ratings. Additionally, the precision and granularity of the estimated quality scores in the proposed dataset will be compared with those in existing SDR and HDR IQA datasets.

\bibliographystyle{IEEEbib}
\bibliography{hdr}

\begin{thebibliography}{10}

\bibitem{ITU-R_BT.709}
Recommendation ITU-R BT.709-6,
\newblock ``Parameter values for the {HDTV} standards for production and international programme exchange,'' 2015,
\newblock \url{https://www.itu.int/rec/r-rec-bt.709}.

\bibitem{ITU2020}
{Recommendation ITU-R BT.2020-2},
\newblock ``Parameter values for ultra-high definition television systems for production and international programme exchange,'' 2015,
\newblock \url{https://www.itu.int/rec/R-REC-BT.2020-2-201510-I/en}.

\bibitem{SMPTE2084}
{SMPTE},
\newblock ``{ST 2084:2014} -- high dynamic range electro-optical transfer function of mastering reference displays,'' SMPTE Standard, 2014.

\bibitem{ITU-R_BT2100-3_2025}
{Recommendation ITU-R BT.2100-3},
\newblock ``Image parameter values for high dynamic range television for use in production and international programme exchange,'' 2025,
\newblock \url{https://www.itu.int/rec/R-REC-BT.2100-3-202502-I/en}.

\bibitem{kunkel2016perceptual}
T~Kunkel, S~Daly, S~Miller, and J~Froehlich,
\newblock ``Perceptual design for high dynamic range systems,''
\newblock in {\em High Dynamic Range Video}, pp. 391--430. Elsevier, 2016.

\bibitem{shang2023study}
Zaixi Shang, Joshua~P Ebenezer, Abhinau~K Venkataramanan, Yongjun Wu, Hai Wei, Sriram Sethuraman, and Alan~C Bovik,
\newblock ``A study of subjective and objective quality assessment of {HDR} videos,''
\newblock {\em IEEE Transactions on Image Processing}, vol. 33, pp. 42--57, 2023.

\bibitem{alakuijala2019jpeg}
Jyrki Alakuijala, Ruud Van~Asseldonk, Sami Boukortt, Martin Bruse, Iulia-Maria Comșa, et~al.,
\newblock ``{JPEG~XL} next-generation image compression architecture and coding tools,''
\newblock in {\em Applications of Digital Image Processing XLII}, 2019, vol. 11137, pp. 112--124.

\bibitem{barman2020evaluation}
Nabajeet Barman and Maria~G Martini,
\newblock ``An evaluation of the next-generation image coding standard {AVIF},''
\newblock in {\em 12th International Conference on Quality of Multimedia Experience (QoMEX)}, 2020, pp. 1--4.

\bibitem{artusi2019overview}
Alessandro Artusi, Rafa{\l}~K Mantiuk, Thomas Richter, Philippe Hanhart, Pavel Korshunov, Massimiliano Agostinelli, Arkady Ten, and Touradj Ebrahimi,
\newblock ``Overview and evaluation of the {JPEG~XT HDR} image compression standard,''
\newblock {\em Journal of Real-Time Image Processing}, vol. 16, pp. 413--428, 2019.

\bibitem{alshina2024jpeg}
Elena Alshina, Jo{\~a}o Ascenso, and Touradj Ebrahimi,
\newblock ``{JPEG AI:} the first international standard for image coding based on an end-to-end learning-based approach,''
\newblock {\em IEEE MultiMedia}, vol. 31, no. 4, pp. 60--69, 2024.

\bibitem{P.910}
{Recommendation ITU-T P.910},
\newblock ``Subjective video quality assessment methods for multimedia applications,'' 2008,
\newblock \url{https://www.itu.int/rec/t-rec-p.910/en}.

\bibitem{BT.500}
{Recommendation ITU-R BT.500-15},
\newblock ``Methodologies for the subjective assessment of the quality of television images,'' 2023,
\newblock \url{https://www.itu.int/rec/R-REC-BT.500}.

\bibitem{Narwaria2013}
Manish Narwaria, Myl{\`e}ne P.~Da Silva, Patrick~Le Callet, and R{\'e}mi Pepion,
\newblock ``Tone mapping-based high-dynamic-range image compression: Study of optimization criterion and perceptual quality,''
\newblock {\em Optical Engineering}, vol. 52, no. 10, 2013.

\bibitem{mantel2014comparing}
Claire Mantel, Stefan~Catalin Ferchiu, and S{\o}ren Forchhammer,
\newblock ``Comparing subjective and objective quality assessment of hdr images compressed with jpeg-xt,''
\newblock in {\em 2014 IEEE 16th International Workshop on Multimedia Signal Processing (MMSP)}. IEEE, 2014, pp. 1--6.

\bibitem{korshunov2015subjective}
Pavel Korshunov, Philippe Hanhart, Thomas Richter, Alessandro Artusi, Rafa{\l} Mantiuk, and Touradj Ebrahimi,
\newblock ``Subjective quality assessment database of {HDR} images compressed with {JPEG XT},''
\newblock in {\em 2015 Seventh International Workshop on Quality of Multimedia Experience (QoMEX)}. IEEE, 2015, pp. 1--6.

\bibitem{mikhailiuk2021consolidated}
Aliaksei Mikhailiuk, Mar{\'\i}a P{\'e}rez-Ortiz, Dingcheng Yue, Wilson Suen, and Rafa{\l}~K Mantiuk,
\newblock ``Consolidated dataset and metrics for high-dynamic-range image quality,''
\newblock {\em IEEE Transactions on Multimedia}, vol. 24, pp. 2125--2138, 2021.

\bibitem{liu2024hdrc}
Yue Liu, Zhangkai Ni, Peilin Chen, Shiqi Wang, and Sam Kwong,
\newblock ``{HDRC}: A subjective quality assessment database for compressed high dynamic range image,''
\newblock {\em International Journal of Machine Learning and Cybernetics}, vol. 15, no. 10, pp. 4373--4388, 2024.

\bibitem{AIC2}
{ISO/IEC 29170-2:2015},
\newblock ``{Information technology — Advanced image coding and evaluation — Part 2: Evaluation procedure for nearly lossless coding},'' 2015.

\bibitem{ISOJPEGAIC2025}
{ISO/IEC CD 29170-3},
\newblock ``Information technology — advanced image coding and evaluation — {Part} 3: Subjective quality assessment of high-fidelity images,'' 2025.

\bibitem{testolina2025fine}
Michela Testolina, Mohsen Jenadeleh, Shima Mohammadi, Shaolin Su, João Ascenso, Touradj Ebrahimi, Jon Sneyers, and Dietmar Saupe,
\newblock ``Fine-grained subjective visual quality assessment for high-fidelity compressed images,''
\newblock in {\em 2025 Data Compression Conference (DCC)}, 2025, pp. 123--132.

\bibitem{men2021subjective}
Hui Men, Hanhe Lin, Mohsen Jenadeleh, and Dietmar Saupe,
\newblock ``Subjective image quality assessment with boosted triplet comparisons,''
\newblock {\em IEEE Access}, vol. 9, pp. 138939--138975, 2021.

\bibitem{itu-bt2390-8}
{Report ITU-R BT.2390},
\newblock ``High dynamic range television for production and international programme exchange,'' 2020,
\newblock https://www.itu.int/pub/R-REP-BT.2390/.

\bibitem{hdr_explained}
Eric Chan,
\newblock ``High dynamic range explained,'' Adobe Blog, 2023,
\newblock \url{https://blog.adobe.com/en/publish/2023/10/10/hdr-explained}.

\bibitem{jenadeleh2023relaxed}
Mohsen Jenadeleh, Johannes Zagermann, Harald Reiterer, Ulf-Dietrich Reips, Raouf Hamzaoui, and Dietmar Saupe,
\newblock ``Relaxed forced choice improves performance of visual quality assessment methods,''
\newblock in {\em 15th International Conference on Quality of Multimedia Experience (QoMEX)}, 2023, pp. 37--42.

\bibitem{ITU-R_BT2246-8_2023}
{Report ITU-R BT.2246-8},
\newblock ``The present state of ultra-high definition television,'' 2023,
\newblock \url{https://www.itu.int/pub/R-REP-BT.2246/}.

\bibitem{meng1992comparing}
Xiao-Li Meng, Robert Rosenthal, and Donald~B Rubin,
\newblock ``Comparing correlated correlation coefficients.,''
\newblock {\em Psychological Bulletin}, vol. 111, no. 1, pp. 172, 1992.

\bibitem{cvvdp}
Rafal~K Mantiuk, Param Hanji, Maliha Ashraf, Yuta Asano, and Alexandre Chapiro,
\newblock ``{ColorVideoVDP}: A visual difference predictor for image, video and display distortions,''
\newblock {\em ACM Transactions on Graphics (TOG)}, vol. 43, no. 4, pp. 1--20, 2024.

\bibitem{hdr-vdp-2}
Manish Narwaria, Rafal~K Mantiuk, Mattheiu~Perreira Da~Silva, and Patrick Le~Callet,
\newblock ``{HDR-VDP-2.2}: A calibrated method for objective quality prediction of high-dynamic range and standard images,''
\newblock {\em Journal of Electronic Imaging}, vol. 24, no. 1, pp. 010501--010501, 2015.

\bibitem{hdr-vdp-3}
Rafa{\l}~K. Mantiuk, Dounia Hammou, and Param Hanji,
\newblock ``{HDR-VDP-3}: A multi-metric for predicting image differences, quality and contrast distortions in high dynamic range and regular content,''
\newblock {\em arXiv:2304.13625}, 2023.

\bibitem{ssim}
Zhou Wang, A.C. Bovik, H.R. Sheikh, and E.P. Simoncelli,
\newblock ``Image quality assessment: from error visibility to structural similarity,''
\newblock {\em IEEE Transactions on Image Processing}, vol. 13, no. 4, pp. 600--612, 2004.

\bibitem{ms-ssim}
Zhou Wang, Eero~P Simoncelli, and Alan~C Bovik,
\newblock ``Multiscale structural similarity for image quality assessment,''
\newblock in {\em The Thirty-Seventh Asilomar Conference on Signals, Systems \& Computers, 2003}. IEEE, 2003, vol.~2, pp. 1398--1402.

\bibitem{psnr-hvs}
Nikolay Ponomarenko, Flavia Silvestri, Karen Egiazarian, Marco Carli, Jaakko Astola, and Vladimir Lukin,
\newblock ``On between-coefficient contrast masking of {DCT} basis functions,''
\newblock in {\em Proceedings of the Third International Workshop on Video Processing and Quality Metrics}. Scottsdale USA, 2007, vol.~4, pp. 1--4.

\bibitem{iw-ssim}
Zhou Wang and Qiang Li,
\newblock ``Information content weighting for perceptual image quality assessment,''
\newblock {\em IEEE Transactions on Image Processing}, vol. 20, no. 5, pp. 1185--1198, 2010.

\bibitem{vmaf}
Zhi Li, Anne Aaron, Ioannis Katsavounidis, Anush Moorthy, and Megha Manohara,
\newblock ``Toward a practical perceptual video quality metric,'' Netflix TechBlog, 2016,
\newblock \url{https://netflixtechblog.com/toward-a-practical-perceptual-video-quality-metric-653f208b9652}.

\bibitem{vmaf-neg}
Zhi Li, Kyle Swanson, Christos Bampis, Lukáš Krasula, and Anne Aaron,
\newblock ``Toward a better quality metric for the video community,'' Netflix TechBlog, 2020,
\newblock \url{https://netflixtechblog.com/toward-a-better-quality-metric-for-the-video-community-7ed94e752a30}.

\bibitem{LGFM}
Yue Liu, Zhangkai Ni, Shiqi Wang, Hanli Wang, and Sam Kwong,
\newblock ``High dynamic range image quality assessment based on frequency disparity,''
\newblock {\em IEEE Transactions on Circuits and Systems for Video Technology}, vol. 33, no. 8, pp. 4435--4440, 2023.

\bibitem{PU21}
Rafa{\l}~K. Mantiuk and Maryam Azimi,
\newblock ``{PU21}: A novel perceptually uniform encoding for adapting existing quality metrics for {HDR},''
\newblock in {\em 2021 Picture Coding Symposium (PCS)}, 2021, pp. 1--5.

\bibitem{butteraugli}
Jyrki Alakuijala,
\newblock ``Butteraugli, a tool for measuring perceived differences between images,'' 2016-2025,
\newblock Original version: \url{https://github.com/google/butteraugli}; current version: \url{https://github.com/libjxl/libjxl/blob/main/lib/jxl/butteraugli/butteraugli.cc}.

\bibitem{ssimulacra1}
Jon Sneyers,
\newblock ``Detecting the psychovisual impact of compression related artifacts using {SSIMULACRA},'' Cloudinary blog, 2017,
\newblock \url{https://cloudinary.com/blog/detecting_the_psychovisual_impact_of_compression_related_artifacts_using_ssimulacra}.

\bibitem{ssimulacra2}
Jon Sneyers,
\newblock ``{SSIMULACRA 2 - Structural SIMilarity Unveiling Local And Compression Related Artifacts},'' 2023,
\newblock \url{https://github.com/cloudinary/ssimulacra2}.

\end{thebibliography}

\end{document}